\documentclass[a4paper]{article}
\usepackage{Odyssey2018}
\usepackage{epsfig,amssymb,amsmath}
\usepackage{graphicx}
\usepackage{dsfont}
\usepackage{txfonts}
\usepackage{multirow}
\usepackage{color}

\ninept
\def\mset#1{\varmathbb{#1}}

\def\vec#1{\ensuremath{\bm{{#1}}}}
\def\mat#1{\vec{#1}}
\renewcommand{\vec}[1]{\boldsymbol{\mathrm{#1}}}
\newcommand\norm[1]{\left\lVert#1\right\rVert}

\title{Staircase Network: structural language identification via\\ hierarchical attentive units}

\makeatletter
\def\name#1{\gdef\@name{#1\\}}
\makeatother
\name{{\em Trung Ngo Trong$^1$, Ville  Hautam\"aki$^1$, Kristiina Jokinen$^2$}}

\address{
 $^1$School of Computing, University of Eastern Finland, Finland \\
 $^2$AI Research Center, AIST Tokyo Waterfront, Japan \\
{\small \tt trung@cs.uef.fi, villeh@cs.uef.fi, kristiina.jokinen@aist.go.jp} }
\begin{document}
\maketitle


\begin{abstract}
Language recognition system is typically trained directly to optimize classification error on the target language labels, without using the external, or meta-information in the estimation of the model parameters. However labels are not independent of each other, there is a dependency enforced by, for example, the language family, which affects negatively on classification. The other external information sources (e.g. audio encoding, telephony or video speech) can also decrease classification accuracy.
In this paper, we attempt to solve these issues by constructing a deep hierarchical neural network, where different levels of meta-information are encapsulated by attentive prediction units and also embedded into the training progress.
The proposed method learns auxiliary tasks to obtain robust internal representation and to construct a variant of attentive units within the hierarchical model.
The final result is the structural prediction of the target language and a closely related language family. The algorithm reflects a ``\textit{staircase}'' way of learning in both its architecture and training, advancing from the fundamental audio encoding to the language family level and finally to the target language level. This process not only improves generalization but also tackles the issues of imbalanced class priors and channel variability in the deep neural network model. Our experimental findings show that the proposed architecture outperforms the state-of-the-art i-vector approaches on both small and big language corpora by a significant margin.
\end{abstract}

\section{Introduction}
\label{sec:intro}
The availability of large corpora in speech processing has been one of the major driving forces advancing speech technologies \cite{KongAik:Interspeech2016:SharedView,deepspeech2}. This has allowed recent state-of-the-art systems to obtain impressive performance in recognizing spoken languages \cite{KongAik:Interspeech2016:SharedView,Ville:ivectorattribute,deepspeakerlang}. However, the focus of the recent NIST Language Recognition Evaluation (LRE) has been shifted to closely related languages in LRE'15 \cite{nistlre15}, then to a mixture of multiple language groups and their dialects in LRE17 \cite{nistlre17}.
As for the challenges of recognizing more confusing groups of languages, most of the developments in the spoken language recognition field have concentrated on two directions: fine-tuning existing algorithms on the new corpora, and applying recent advances in the field of machine learning in order to obtain improvements in the recognition accuracy \cite{KongAik:Interspeech2016:SharedView,10.1371/journal.pone.0146917}. However, the new corpora also provides an interesting test case for more detailed analyses, since the target languages are grouped into language families, and two different types of audio recordings are also introduced.

An effect similar to the language family effect can also be seen in a much smaller setting when we, for instance, look at the regional dialects of the spoken in the Northern parts of Norway, Finland, and Sweden. The Sami languages cover a large geographical area in  Northern Europe, and North Sami is the biggest group with about 20000 speakers. It is an official language, but the speakers are at least bilingual and also speak the dominant language of their area, i.e. Finnish, Norwegian, or Swedish. In \cite{Jokinen+2016}, we have previously shown that the North Sami regional dialects are confounded by multiple hidden factors which significantly influence the performance of an i-vector based system.
One of the factors seems to be the variation among the speakers caused by the majority language: Finnish in Finland and Norwegian in Norway. Thus, in the same way as the language family effect is observed in LRE'17, we can observe the majority language effect on the regional dialect recognition in the North Sami case. It should be noticed that North Sami and Finnish belong to the Finno-Ugric language family while Norwegian is an Indo-European language, so a language family effect might play a role here, too; however, the regional variation in North Sami is likely to be caused by the country borders and minority-majority language distinction which have reinforced the North Sami speakers to accommodate their North Sami speaking according to their second language, i.e. the majority language Finnish or Norwegian.

In this paper, we aim at a direct algorithmic approach to tackle the problem of explicitly modeling these confounding factors by addressing the challenges seen in the  language family and the audio encoding effects. Since the target languages in LRE17 are grouped into language families \cite{nistlre17}, we can use them to construct a shared representation of two or more languages. By leveraging a simpler auxiliary task, i.e. identifying the language family, the recognizer efficiently learns how to learn the language characteristic. Similar approaches using multiple domains or tasks to optimize task-specific representation have shown significant improvement on the original objective \cite{MultiWay:Multilingual:SharedAttention,multitask_hard,multitask_soft}. In designing multi-task systems, it is important  to avoid focusing on unwanted variability, since this could drive the attention of the algorithm away from its main task \cite{DBLP:journals/corr/Ruder17a}. Conventional approaches to multi-task learning implicitly model these factors, and are heavily dependent on the assumption that learning them allows to model to learn beneficial shared representations \cite{DBLP:journals/corr/Ruder17a}. Thus, we propose a solution for learning the hierarchy of tasks as well as attending to relevant patterns evolved from specific connection among tasks.

Different characteristics of audio recording and utterance are mixed together in a  non-linear way, so it remains a challenge to perform speech processing task over very different types of audio collections. As in LRE17, we have a new audio encoding format, which has shown to degrade the LID performance.
Prior studies have mainly focused on data augmentation, feature post-processing, score adaptation and transfer learning \cite{KongAik:Interspeech2016:SharedView,transfer_lid} with a certain exploration on integrating this task into a learnable classifier. As a result, we present a technique for embedding the effect of audio encoding variation into the training process of a deep neural network.

\section{Staircase Methods}
\label{sec:methods}

In this section, we present our ``staircase'' strategy for language learning via deep representation.
We first introduce the necessary notation and give an overview of our method as well as recent advances in the field.
We then delve into the details of the methods.

\subsection{Background and notation}
Our task is to recognize language $\vec{y}_{i \in \mset{L}}$ from a set $\mset{L}$ different languages. Subsequently, the languages are grouped into language families, $\mset{F}$.
Moreover, the utterances were recorded in different sessions by a multitude of recording devices, so we obtain a set of encodings $\mset{E}$, such as VAST and MLS14 in NIST LRE17 which will be introduced in Section~\ref{sec:exp_lre17}.
In general, all of the above information is provided in the  training set $\mset{D}^t=\{(x_i, y_{i,j,k})|i\in\mset{L};j\in\mset{F};k\in\mset{E}\}_{i=1}^{N^t}$. Also, we have a set of unlabeled data for evaluation $\mset{D}^e=\{(x_i)\}_{i=1}^{N^e}$. The goal is to optimize the language classifier by leveraging a series of auxiliary tasks. This process can be seen as collaborative feature learning or multi-task learning since we try to obtain the objective by using a single representation (i.e. deep neural network).

\begin{figure*}[t]
  \centering
  \includegraphics[width=1.02\linewidth]{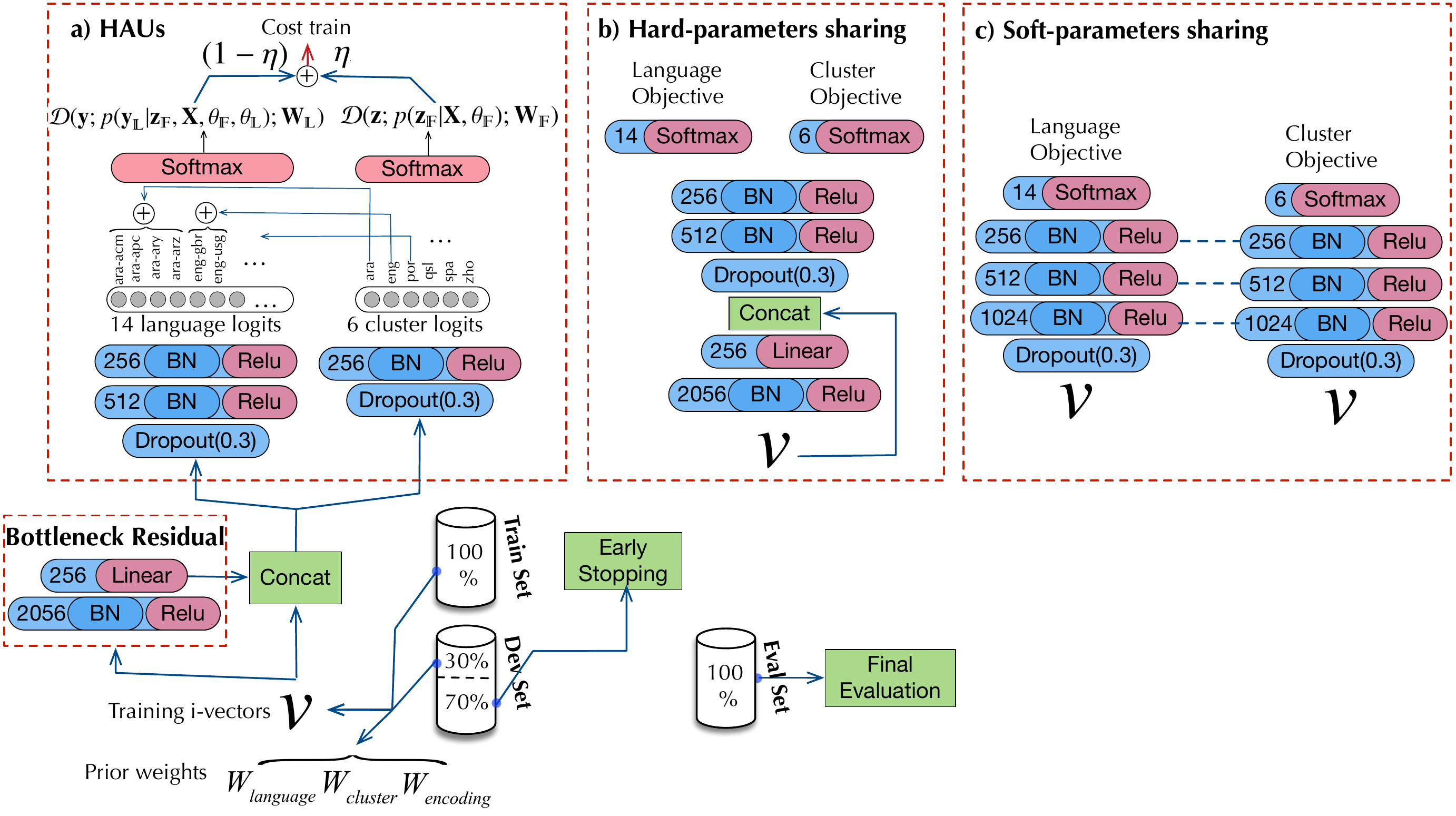}
  \caption{Illustration of our strategy fine-tuned for the NIST LRE17, also present the design of, hierarchical attentive units (HAUs) in comparison to hard parameters sharing and soft parameters sharing approaches.}
  \label{fig:all_system}
\end{figure*}

In the context of deep learning, multi-task learning can be achieved by optimizing the same set of parameters under influence of multiple objectives. There exists two common strategies for sharing those parameters.
The hard parameter sharing method \cite{multitask_hard}, or the multi-head network \cite{alpha_zero} shares the hidden layers between all tasks, while creating multiple ``heads'' (i.e. output layers) specifically for each task as illustrated by Fig~\ref{fig:all_system} (b).
As suggested in \cite{multitask_hard}, hard parameter sharing can greatly reduce the risk of overfitting, and at the same time learning beneficial representation from multiple tasks simultaneously.
The soft parameter sharing have been proposed in \cite{multitask_soft}, and the design is described in Fig~\ref{fig:all_system} (c). The algorithm maintains multiple instances of the same network for each task. In order to enforce the learning of shared representation, additional regularization is added to minimize the distance of the coordinated parameters among the networks. As a result, the objective becomes overly complicated as the number of tasks exceeds three.

Since both of these techniques implicitly model the multi-task case, meaning that the algorithm is heavily based on the assumption that anything learned from the extra tasks will benefit the main task. As a result, they deliberately ignore the relation among the targets within each task, and the structural relation between the auxiliary tasks and the main task. They introduce the risk of modeling irrelevant pattern as well as lessen the important role of the main objective. Our present work explores solutions to these problems in two directions:
\begin{enumerate}
  \item Predicting relevant characteristics (i.e. language family) to the main task (Section~\ref{sec:haus}).
  \item Jointly learning the closely related task (i.e. utterances' encoding) (Section~\ref{sec:bayes_ce}).
\end{enumerate}

\subsection{Hierarchical attentive units (HAUs)}
\label{sec:haus}

The intuition behind learning the language families is that if a language belongs to e.g. Arabic languages, two things must be enforced:
\begin{itemize}
  \item it must not belong to any other family (e.g. Chinese, Iberian, ...),
  \item the language could only be one of the given languages within the set of all Arabic languages.
\end{itemize}
However, modeling such correlation requires the network performing reasoning, for instance, if the language is likely to be Arabic, then, it should prioritize predicting 4 different Arabic dialects (e.g. Egyptian, Iraqi, Levantine, Maghrebi).
Implementing an ``if-then-else'' logic in a differentiable system while preserving the end-to-end design of deep learning is problematic.

Inspired by the approach \cite{Morin05hierarchicalprobabilistic}, we propose a hierarchical prediction system that is able to attend to a subset of predictive neurons during inference. We tie the connection between the language and its family by adding the families' logit values to the appropriate languages (Fig~\ref{fig:all_system} (a)). These values weigh up the relevant languages within the chosen group, and weighs down all the other languages.

In order to realize this idea, we design two linear output layers: $\mathbf{f}_{\mset{F}}$ for the language families, and $\mathbf{f}_{\mset{L}}$ for the target languages. Both layers use softmax activations to predict the language family and target language accordingly for each utterance. The first network procedures the logit values by $\ell_{\mset{F}}$:
\[
    p(\vec{z}_{\mset{F}}|\mat{X},\theta_{\mset{F}}) = \mathrm{softmax}(\ell_{\mset{F}}) \\
\]
where $\vec{z}_{\mset{F}}$ refers to the predicted probabilities for each language group and $\theta_{\mset{F}}$ is the set of all parameters used for $\mathbf{f}_{\mset{F}}$. Subsequently, we weigh the language prediction logits $\ell_{\mset{L}}$ by the following formula
\begin{equation}
    \label{eq:haus}
    p(\vec{y}_{\mset{L}}|\vec{z}_{\mset{F}},\mat{X},\theta_{\mset{F}},\theta_{\mset{L}}) =
    \mathrm{softmax}\Big(
    \bigoplus_{j\in\mset{F}} (\ell_{\mset{L}}^{(i) \forall i \in \mset{L}; F(i) = j} +
                              \ell_{\mset{F}}^{(j)})
                    \Big),
\end{equation}
where $\bigoplus$ is a concatenating operator for a sequence of vector, $F(.)$ is the operator extracting the language family of a given language, and $\theta_{\mset{L}}$ is the set of all parameters used for $\mathbf{f}_{\mset{L}}$.
Eq.~(\ref{eq:haus}) can be explained as follows: ``for each language family, add its logit value to the appropriate logit units of all the languages of the family in the language classifier''.

As it is important to penalize the network more for making a wrong decision on an easy task, we use $\eta$ to balance the contribution of the two objectives: language grouping and language recognizing. The final training cost is
\begin{equation}
\begin{aligned}
    \Im =&
  \eta \cdot \mathcal{D}(\vec{z}; p(\vec{z}_{\mset{F}}|\mat{X},\theta_{\mset{F}}); \vec{W}_{\mset{F}}) + \\
  &
  (1 - \eta) \cdot \mathcal{D}(\vec{y}; p(\vec{y}_{\mset{L}}|\vec{z}_{\mset{F}},\mat{X},\theta_{\mset{F}},\theta_{\mset{L}}); \vec{W}_{\mset{L}}),
\end{aligned}
\end{equation}
where $\mathcal{D}(\cdot)$ is the discriminative loss function, $\vec{W}_{(.)}$ is a specific weight for each training example initialized to be $1.0$ by default. The $\eta$ value should be greater than $0.5$, which emphasizes that a bigger weight is put on clustering a given language to the right language family.
It should be noted that the labels of language family are only provided during training, and the network is \textit{self-attentive}, i.e. using its own group prediction for weighting relevant languages during testing.

Additionally, $\theta_{\mset{F}}$ and $\theta_{\mset{L}}$ could be an overlapping set of parameters. However, our experiments have shown that a deeper architecture should be used for more complicated task (i.e. predicting the actual language), and a  shallower network can be used for simpler task (i.e. predicting language family). As a result, we propose the design Fig~\ref{fig:all_system} (a) for language identification on LRE17.

\subsection{Cost adaptive objective}
\label{sec:bayes_ce}

After the training, we observed high variation of the performance among different utterance encodings. The issue can be traced to the imbalance in the utterance distribution between encodings, which has a strong negative impact on the network generalization performance \cite{deeplang,cnnimbalanced}. Specifically, each training step can drive the network to a different sub-optimal solution created by the dominant classes \cite{deeplang}.
Since deep learning, in general, can be seen as an automatic feature learning algorithm \cite{deeplearning}, the network should adapt its representation for modeling the language pattern in all encodings.
We introduce a cost-sensitive objective function to let the network learn better adaption among the encodings. The idea here is that if the objective can balance the importance of each encoding internally, the network will be able to focus on all classes equally, and as a result, it could learn crucial language characteristics from all encodings.

The modified version of cross-entropy \cite{cnnimbalanced} (Bayesian cross-entropy - BCE) takes into account the prior distribution of the training set, and scales the loss value appropriately for each class
\[
\mathcal{D}(\theta|(X,y)) = -\frac{1}{K n} \sum_{i=1}^n y_i *
                            \frac{\mathrm{log} (f(x_i, \theta))}
                            {\mathrm{Pr}(y_i)},
\]
where $n$ is the number of training examples, $K$ is the number of classes, and $\mathrm{Pr}(y_i)$ is the prior probability of class $y_i$. We then estimate the prior probabilities based on the number of samples from each class in the training set, and use these priors for calculating the weights for each example during training:
\begin{equation}
\begin{aligned}
    Pr_i &=  \frac{n_i}{\sum_j^Cn_j}, \\
    w_i &= \frac{\underset{j=1..C}{\max}(Pr_j)}{Pr_i},
\end{aligned}
\end{equation}
where $C$ is the number of classes (in LRE17: 14 for language, 6 for language family and 2 for encoding), and $n_i$ is the number of classes within the training set. The $w_i$ is the weight for $i^{th}$ classes, which is the inverted ratio of each class compared to the most dominant label. However, this value could be huge due to some extreme cases, and consequently, makes the network over-concentrated on the minority classes. We rescale the weights using
\begin{equation}
\begin{aligned}
    w_{\min} &= \underset{j=1..C}{\min}(w_j), \\
    w_{\max} &= \underset{j=1..C}{\max}(w_j), \\
    w_i &= \frac{(x_{\max} - x_{\min}) * (w_i - w_{\min})}{w_{\max} - w_{\min}} + w_{\min},
\end{aligned}
\end{equation}
where $[x_{\min}, x_{\max}]$ are the lower and upper bounds of the weights and are chosen experimentally.  We selected the following values $[0.1, 8.0]$.
Using the strategy, we estimate the weights for different audio encoding (\textit{VAST} and \textit{MLS14})-$W_{\mathrm{encoding}}$, language-$W_{\mathrm{language}}$, family-$W_{\mathrm{cluster}}$ (Fig~\ref{fig:all_system}-bottom). The weight for each task (i.e. recognizing language and language group) is given as the combination of task-specific prior weights and the channel prior weights, as follow
\begin{equation}
  \begin{cases}
    \vec{W}_{\mset{F}} = \vec{W}_{\mathrm{encoding}} + \vec{W}_{\mathrm{cluster}} \\
    \vec{W}_{\mset{L}} = \vec{W}_{\mathrm{encoding}} + \vec{W}_{\mathrm{language}} \\
  \end{cases}
\end{equation}
As a result, the training weight for each example is selected according to each task. Because the difference between encoding is significant, the network matches the weight of each sample to its appropriate encoding without the need for external labels.
It has been shown in our experiments Sec~\ref{sec:exp_lre17} that modeling encoding directly causes performance deficit. Since recognizing encoding is not as closely related task as identifying language family, the approach saturates the main objective.

\section{Experiments}
In this section we validate the presented methods in Section~\ref{sec:methods} by two experiments:
\begin{itemize}
  \item First, we leverage the two extensions for improving the classifier backend of LID system on the large NIST corpus.
  \item Second, we extend the applicability of the method to boost the performance of end-to-end dialect recognition for under-resourced languages using a North Sami corpus. The experiment highlights the crucial role of modeling confounding factors in case the external information is limited and unavailable.
\end{itemize}

\subsection{Improving backend on NIST LRE17}
\label{sec:exp_lre17}

\subsubsection{NIST LRE17 corpus}

The 2017 NIST language recognition evaluation (LRE17) is the most recent effort to advance the research in LID. The challenge as described in the evaluation plan \cite{nistlre17} builds on the history of the LRE campaigns, and it shares many features with the previous challenges. However, there are two major differences that propose challenges to the community.

\begin{itemize}
  \item The inclusion of \textit{VAST} utterances in the development set and the evaluation set. The audio were extracted from video data using encoding and channel variation which were very different from the traditional telephony speech in \textit{MLS14} corpus.
  \item Normalized cross-entropy ($C_{norm}$) is used as performance metrics. The evaluation process calculates $C_{norm}$ for each language under two assumed prior probabilities $P_{true}=0.5$ and $P_{true}=1.0$. The final score is the average of all these values and provided in our experiments.
\end{itemize}

Our strategy for splitting the dataset is presented in Fig~\ref{fig:all_system}. The goal is to evaluate the ability of each algorithm in domain adaptation, i.e. match the performance on both \textit{MLS14} and \textit{VAST} utterances.
As a result, we limited the amount of VAST exposed to the training algorithm by randomly selecting only 30\% of development set for the training set. The remaining part, named as \textit{validation set}, is then for early-stopping, hyper-parameters tunning, validation and alternative evaluation. We remind that the evaluation data is untouched and only presented during scoring phase. There are 17425 files for training, 2440 files for validation and 25449 files for evaluation.

\subsubsection{Bottleneck features based i-vector}
\label{sec:bnf_ivec}

We use state-of-the-art bottleneck-based i-vector extractor \cite{KongAik:Interspeech2016:SharedView, deepspeakerlang} as the basis of our LID systems. A bottleneck DNN was trained using the 13-dimensional MFCC features with concatenated delta and delta-delta coefficients extracted from the \textit{Switchboard-1} and \textit{Fisher} corpora ($\approx$ 2000 hours). The features were then applied per utterance mean and variance normalization, then stacked by 10 past and future frames to form 21 contextual acoustic input. The seven hidden layers have 2048 units each and the bottleneck layer, which has 80 units, is placed 2 layers before the output layer. We use ReLU activation followed by a re-normalization that scales the activations RMSE to 1.0, with the exception of the bottleneck layer, which only has re-normalization nonlinearity. The output layer has 8700 targets corresponding to a set of 8700 senones obtained from the baseline speaker-independent GMM-HMM system. The 80-dimensional bottleneck features are used for extracting the i-vectors. An energy-based voice activity detection (VAD) technique was applied to the raw bottleneck features to exclude the silence frames. The voiced frames were then used to train a universal background model (UBM)  with 2048 Gaussians with diagonal covariances. The diagonal UBM was used to train the total variability matrix and extract the 400-dimensional i-vectors.

{\em Within-class covariance normalization} (WCCN) \cite{wccn} was used to compensate unwanted intra-class variation in the total variability space~\cite{Behravan2014spec}. We then use {\em linear discriminant analysis} (LDA) to project the i-vectors onto a sub-space where inter-dialect variability is maximized and intra-dialect variability is minimized. This post-processing is repeated for all experiments with the backend to ensure the comparable results.

\subsubsection{Baseline backend classifiers}

We present the three most common i-vector backends: cosine scoring, support vector machine (SVM) \cite{Cristianini:1999:ISV:345662}, and multi-class logistic regression (MCLR), together with various configuration for the deep classifier.

Given two i-vectors $\vec{w}_{\mathrm{test}}$ and $\vec{w}_{\mathrm{target}}^l$ for the language $l$, cosine similarity score $t$ is computed as follows:
\begin{equation}
t = \frac{ \vec{\hat{w}}^T_{\mathrm{test}} \vec{\hat{w}}^d_{\mathrm{target}} }
         { \norm{\vec{\hat{w}}_{\mathrm{test}}} \norm{\vec{\hat{w}}^d_{\mathrm{target}}} }
\end{equation}
The representation of $l^{th}$ target language, $\vec{\hat{w}}^d_{\mathrm{target}}$, can be simply obtained by taking the average of the corresponding i-vectors in the training set. This score a calculated for all target languages, and we identify the language by the one with the highest degree of similarity.

SVM as a language classifier involves two issues: the algorithm must handle multi-classification task and deal with the non-linearity of speech and language representation. As a result, we train a multi-class SVM according to a one-vs-one scheme. We empirically selected radial basis function (RBF) kernel after it outperformed other options (linear, polynomial and sigmoid kernel).
The optimization process relies upon a maximum margin concept, the focus of SVM is to model the boundary between the positive and the negative classes as opposed to a traditional density estimation approaches.

Direct discriminative training of the languages can be performed via MCLR. The backend leverages gradient descent to minimize the log-likelihood of its predicted probability values to the target language distribution in the training set. We also apply the most recent advances in training deep neural networks to obtain better MCLR solutions. We use mini-batch training, and the adaptive learning rate method (\textit{ADADELTA}) \cite{adadelta} is used instead of the conventional stochastic gradient descent. Implicit regularization \cite{UnderstandDL:Generalization} via early stopping \cite{earlystopping} is used to increase generalizability. The criterion is \textit{generalization loss} \cite{earlystopping} and the learning rate decay is $0.96$. Whenever the network drops its validating score we rollback the parameters to the best checkpoint.

\subsubsection{Results}

Table~\ref{tab:other_backends} compares the performance of different classifiers on the LRE17 corpus. All hyper-parameters of each method are fine-tuned using validation set. Our proposed approach (i.e. HAUs Section~\ref{sec:haus}) outperforms other methods by a significant margin. The performance on both validation set and evaluation set are closely matched. However, HAUs performed better on validation set than on the evaluation set. This could be a sign of overfitting or that latent representation has been learned effectively.  Since the performance on the evaluation set has also improved, the second explanation is more plausible.

\begin{table}[th]
  \caption{Comparison, in terms of the actual cost ($C_{\mathrm{primary}}$), between different backends on the LRE17 corpus.}
  \label{tab:other_backends}
  \centering
\resizebox{0.8\linewidth}{!}{
  \begin{tabular}{c|c|c}
    \hline
    Methods  &  Validation set & Evaluation set \\
    \hline
    Cosine scoring               & 22.15          & 21.48 \\
    SVM                          & 17.57          & 17.55 \\
    MCLR                         & 18.63          & 18.52 \\
    HAUs (Sec.~\ref{sec:methods}) & \textbf{16.29} & \textbf{16.88}
  \end{tabular}}
\end{table}

\begin{table}[th]
  \caption{Comparison, in terms of the actual cost ($C_{\mathrm{primary}}$), between other deep neural network architecture variants on the LRE17}
  \label{tab:other_variants}
  \centering
\resizebox{1.02\linewidth}{!}{
  \begin{tabular}{c|c|c|c|c|c|c}
    \hline
    \multirow{2}{*}{Methods}   &  \multicolumn{3}{c|}{Validation set} & \multicolumn{3}{c}{Evaluation set} \\
    \cline{2-7}
                                  &  mls14 & vast  & avg   & mls14 & vast  & avg \\
    \hline
    Single task          & 15.21           & 21.01          & 16.77          & 17.68          & 19.61 & 18.05 \\
    Multitask-hard       & \textbf{15.01}  & 19.38          & 16.19          & 17.59          & \textbf{18.37} & 17.66 \\
    Multitask-soft       & 15.25           & 21.02          & 17.06          & 18.37          & 19.82 & 18.64 \\
    No HAUs              & 15.19           & \textbf{18.89} & \textbf{16.01} & 17.61          & 18.76 & 17.80 \\
    No BCE          & 18.70           & 23.82          & 20.06          & 19.65          & 23.25 & 20.08 \\
    HAUs (Sec.~\ref{sec:methods}) & 15.12           & 19.88          & 16.29          & \textbf{16.65} & 18.35 & \textbf{16.88} \\
  \end{tabular}}
\end{table}

Table~\ref{tab:other_variants} delves deeper into the question "why does HAUs work?". First, we validate the performance of a fine-tuned single task network instance. The performance gap between validation and evaluation set shows a strong sign of overfitting. The same observation applies to both multi-task systems with either hard or soft parameter sharing, and our design with HAUs objective removed. In the multi-task learning, the loss surface dimensions are exponentially increased with the number of tasks, and as a result, the searching for the state of equilibrium becomes difficult without sufficient amount of training data. Subsequently, the learning of the added tasks could diminish learned representation that benefits the main task. On the other hand, removing HAUs extension while keeping the adaptive objective forces the algorithm to improve VAST on the validation set, but fails to transfer this improvement to the evaluation set.

Finally, omitting the proposed objective (i.e. ``No BCE'') significantly degrades the performance due to the negative impact of the imbalance in the training set proportions. We see that the performance on VAST is degraded significantly, whereas HAUs keeps the model generalizing well and matching similar performance on both validation and evaluation set. We can see that applying both extensions from Section~\ref{sec:methods} yields the highest performance single system for the BNF i-vector input.

\begin{table}[th]
  \caption{The performance of HAUs on evaluation set given different $\eta$ values (Eq.~\ref{eq:haus}).}
  \label{tab:other_eta}
  \centering
  \begin{tabular}{c|c|c|c|c|c|c}
    \hline
    $\eta$        & 0.3   &  0.4  & 0.5   & \textbf{0.6}   & 0.7   & 0.8 \\
    \hline
    $C_{primary}$ & 16.98 & 16.98 & 16.90 & \textbf{16.88} & 16.96 & 17.18
  \end{tabular}
\end{table}

Although HAUs introduces another hyper-parameters $\eta$ (Eq.~\ref{eq:haus}), Table~\ref{tab:other_eta} highlights the robustness of HAUs. With all selected values of $\eta$, HAUs still outperforms other backend classifiers and variants from Table~\ref{tab:other_backends} and Table~\ref{tab:other_variants}. The best choice for $\eta$ ranges from 0.5 to 0.7 which matches our intuition and suggestion in Sec.~\ref{sec:haus}, that the easier task should be penalized more during the training process.

\subsubsection{Further analysis}

\begin{figure*}[t]
  \centering
  \includegraphics[width=1\linewidth]{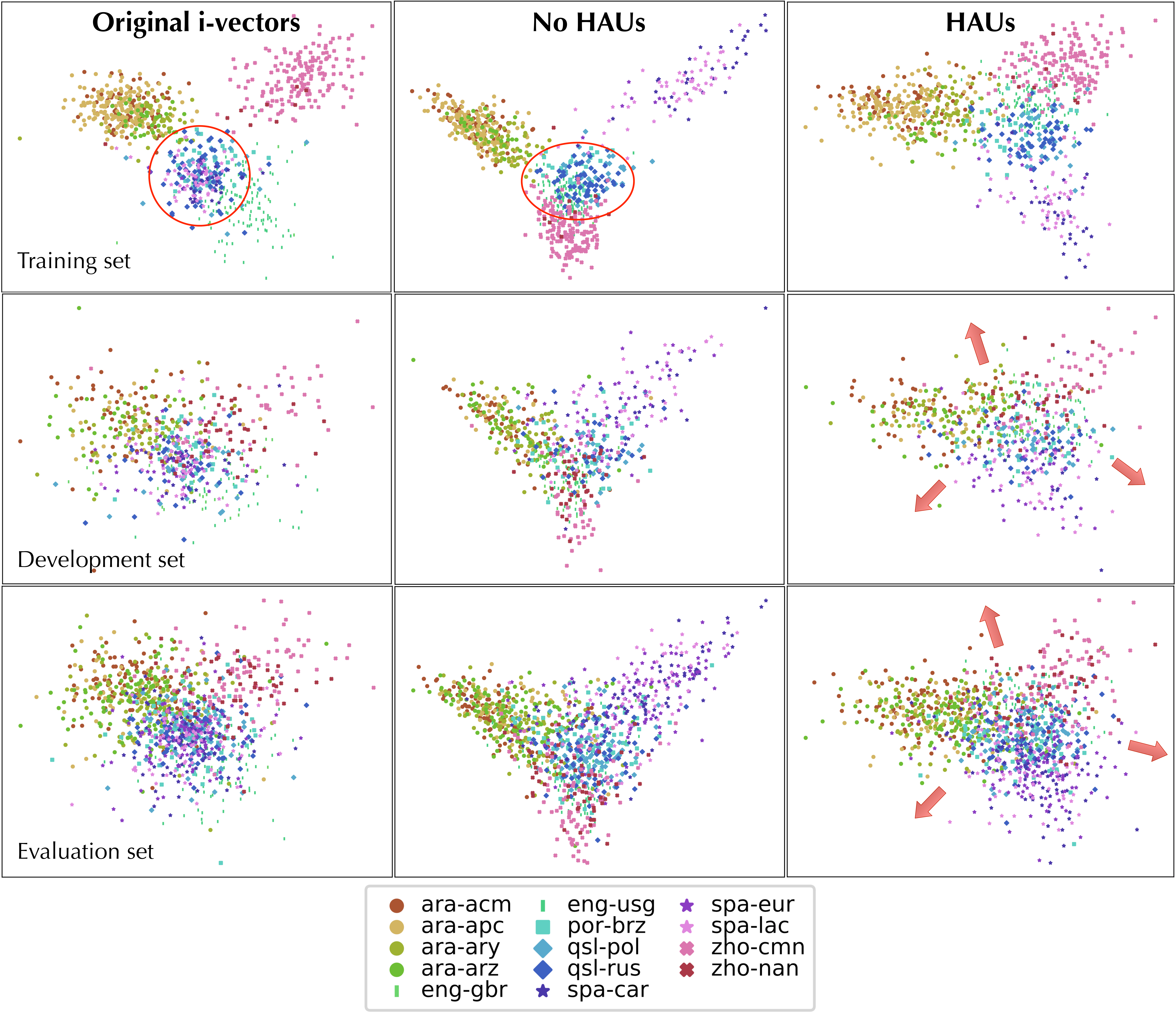}
  \caption{Comparison of learnt representation on training, validation and evaluation sets among the three learnt representations: original i-vectors, last hidden activation from the ``No HAUs'' network, last hidden activation from the ``HAUs'' network.}
  \label{fig:pca_lda}
\end{figure*}

Fig~\ref{fig:pca_lda} illustrates the differences on the learned representations among the three datasets. The plots are acquired by training PCA on the training set followed by LDA, then applying the trained model on validation and evaluation sets. All data points from the same language family have the same marker style, and each language is represented by different colors. While the first and second representations are highly separable by language family on the training set, they encapsulate strong confusion region marked by the red circle. The region contains data points densely populated within a small area. It is also notable that PCA-LDA failed to generalize to the validation and evaluation sets since the samples are all mixed together regardless of their language group.

Deeper classifier (i.e. ``No HAUs'') is able to separate the language family within its representation, but the confusion remains high in the center. However, the learned representation by HAUs is spread over a wider area (directions denoted by red arrows), while the algorithm maintains the same margin among language families. This can be understood as HAUs having been able to learn a richer language family space. As a result, it is able to reduce the confusion by not grouping the languages into big clusters only, but also into mini-clusters.

\subsection{End-to-end approach for under-resourced corpora}

We apply Bayesian cross-entropy objective and HAUs for an end-to-end approach to dialect recognition on under-resourced corpora. However, the objective heavily relies on the assumption that our training set encapsulates the same distribution as the population. The assumption is unlikely to be met in the scenarios of a small dataset, where we only observe an insufficient small number of samples. As a result, we use mini batch statistics to aggregate the prior probability of each class, $\mathrm{Pr}(y_c) = \sum_i^n \mathds{1}(y_i = c) / n$. T This approach has been proven to stabilize the gradients and leads to better results in our experiments.

\subsubsection{DigiSami corpora}
The DigiSami corpora \cite{JokinenKristiina14,Jokinen2017} contain two North Sami corpora recorded in the locations of three villages in Finland: Utsjoki, Inari and Ivalo, and two villages in Norway: Kautokeino and Karasjoki. The participants were invited to take part in two different tasks: discussion and reading aloud existing Wikipedia texts, and taking part in a free conversation. Table~\ref{tab:dist} shows the distribution of the two collected corpora. There are 10 men and 18 women. Their age ranges from 16 to 65 years: 17 were 16-21 years old, five 30-44 years old, and six 49-65 years old. The participants are native speakers of North Sami and are also bilingual with Finnish or Norwegian depending on their home area.

\begin{table}[th]
  \caption{The two corpora statistics, the numbers in the brackets indicate number of new speakers in Conversational corpus.}
  \label{tab:dist}
  \centering
\resizebox{0.8\linewidth}{!}{
  \begin{tabular}{c|c|c|c|c}
    \hline
    \multirow{2}{*}{Dialects}  &  \multicolumn{2}{c|}{Read speech} & \multicolumn{2}{c}{Conversational speech} \\
    \cline{2-5}
                               &  \#Spk & Hour & \#Spk & Hour \\
    \hline
    Kautokeino & 4 & 1.03 & - & - \\
    Karasjoki  & 6 & 0.72 & 6(1) & 1.5 \\
    Ivalo      & 6 & 0.72 & 7(1) & 0.72 \\
    Utsjoki    & 5 & 1.07 & 6(1) & 1.03 \\
    Inari      & 4 & 0.73 & - & - \\
    \hline
    \textbf{Total}      & \textbf{25} & \textbf{3.26} & \textbf{19} & \textbf{4.28} \\
  \end{tabular}}
\end{table}

\subsubsection{Baseline systems}
We use the state-of-the-art i-vector approach as described in Section~\ref{sec:bnf_ivec}. To compensate for the lack of sufficient development data, we first trained the UBM and the T-matrix using the Finnish language \emph{PERSO} corpus \cite{PERSO} corresponding to 30916 utterances in total. The backend is replaced by {\em heteroscedastic linear discriminant analysis} (HLDA)~\cite{Behravan2014spec} and the cosine similarity is used subsequently.

We use the combination of CNN, LSTM and feedforward neural network as described in \cite{deeplang}. The network design is specifically fine-tuned for our task and re-scaled to match the size of the North Sami corpora. The fine-tuning strategy is gradually augmenting the network architectures by increasing the number of layers following by a number of hidden units \cite{deeplang}. The network consists of two convolutional layers with 32 and 64 features map of $3\times3$ kernel, then follows by a bidirectional LSTM of 256 units each. Finally, we include a projection layer of 512 units and an output layer.

\subsubsection{Results and analysis}
As discussed in \cite{iwsds18_sami}, all the experiments are repeated three times to minimize the effect of random initialization on a small dataset, the final reported numbers are the mean and the standard deviation of these experiments. The whole conversational corpus is left out for evaluation since we want to validate the generalization ability under different session and recording. We use ``leave-one-speaker-out'' (LOSO) strategy \cite{Behravan2014spec}, i.e. one speaker from each dialect is randomly selected for validation process, and another speaker is selected for testing. The remaining set of speakers is used for training our classifier. As the scheme used in \cite{Jokinen+2016}, the results are comparable to previous i-vector approach \cite{Jokinen+2016}. It is notable that each repetition is done with a totally different set of speakers in order to preserve the generalizability of the experiments.

Table~\ref{tab:evaluation} emphasizes the importance of adaptive objectives and HAUs in improving classifier performance in case of under-resourced languages. Moreover, our best network significantly outperforms the i-vector system.

\begin{table}[th]
  \caption{Results reported in term of $C_{\mathrm{avg}} \times 100$}
  \label{tab:evaluation}
  \centering
\resizebox{0.77\linewidth}{!}{
  \begin{tabular}{c|c|c}
\hline
Methods & Read speech & Conv. speech \\
\hline
cross-entropy           & $11.66 \pm 1.66$        & $13.56 \pm 0.99$  \\
BCE                     & $10.40 \pm 1.03$        & $10.53 \pm 2.43$  \\
BCE + HAUs              & $\boldsymbol{8.73 \pm 3.24}$ & $\boldsymbol{6.81 \pm 0.58}$\\
i-vector \cite{Jokinen+2016} & $17.79$                 & --              \\
  \end{tabular}}
\end{table}

Even though the HAUs objective has been applied to a complicated architecture, it was able to encapsulate the effect of majority language in the network architectures using only supervised data. This opens up the potential of using end-to-end deep learning approach for under-resourced corpora.

\section{Conclusions}
In the present study, we attempted to model jointly, as an auxiliary task, the effect of language family in the audio encoding.
We observed the potential benefit of modeling hierarchical structures implicitly in the learning algorithm, and presented a novel approach to integrate meta-information learning into deep classifier, which outperformed the common approaches to multi-task learning.
We also demonstrated that by modifying the training objective that takes into account the prior information of multiple internal factors will help the learned model to generalize better. Moreover, we studied how the method's applicability can be extended to boost end-to-end dialect recognition for under-resourced languages like North Sami.
As a future work, we plan to investigate how to lessen the effect of supervision in our model, as the supervised model can too easily lack robustness against unforeseen dataset drift, as observed for example in the LRE15 evaluation set. Consequently, we will focus on using semi-supervised learning to explicitly model channel variation.

\section{Acknowledgments}
This research was partly funded by the Academy of Finland (grant \#313970). We gratefully acknowledge the support of NVIDIA Corporation with the donation of the Titan Xp GPU used for this research.

\bibliographystyle{IEEEbib}

\end{document}